%% file: main.tex
\documentclass[10pt,twocolumn,letterpaper]{article}

\usepackage[pagenumbers]{wacv} 

\usepackage{graphicx}
\usepackage{amsmath}
\usepackage{amssymb}
\usepackage{booktabs}
\usepackage{multirow}
\usepackage{siunitx}
\usepackage{pifont}
\usepackage{graphicx}
\usepackage[export]{adjustbox}
\usepackage{cuted}
\usepackage[font=normalfont,labelfont=bf]{caption}
\usepackage[accsupp]{axessibility}

\usepackage[breaklinks,colorlinks,bookmarks=false]{hyperref}

\usepackage[capitalize]{cleveref}
\crefname{section}{Sec.}{Secs.}
\Crefname{section}{Section}{Sections}
\Crefname{table}{Table}{Tables}
\crefname{table}{Tab.}{Tabs.}

\newcommand{\cmark}{\text{\ding{51}}}
\newcommand{\xmark}{\text{\ding{55}}}
\newcommand{\ourmethod}{RendBEV}

\usepackage{tikz}
\usepackage{textcomp}
\newcommand\copyrighttext{%
  \footnotesize \textcopyright 2025 IEEE. Personal use of this material is permitted.
  Permission from IEEE must be obtained for all other uses, in any current or future
  media, including reprinting/republishing this material for advertising or promotional
  purposes, creating new collective works, for resale or redistribution to servers or
  lists, or reuse of any copyrighted component of this work in other works.
  DOI: \href{https://doi.org/10.1109/WACV61041.2025.00062}{10.1109/WACV61041.2025.00062}}

\newcommand\copyrightnotice{%
\begin{tikzpicture}[remember picture,overlay]
\node[anchor=south,yshift=10pt] at (current page.south) {\fbox{\parbox{\dimexpr\textwidth-\fboxsep-\fboxrule\relax}{\copyrighttext}}};
\end{tikzpicture}%
}

\begin{document}

\title{\ourmethod: Semantic Novel View Synthesis\\ for Self-Supervised Bird's Eye  View  Segmentation}

\author{Henrique Piñeiro Monteagudo$^{1,2}$\thanks{Supported by the SMARTHEP project, funded by the European Union’s Horizon 2020 research and innovation programme, call H2020-MSCA-ITN-2020, under Grant Agreement n. 956086}\
    \and
    Leonardo Taccari$^{1}$
    \and
    Aurel Pjetri$^{1,3}$
    \and
    Francesco Sambo$^{1}$
    \and
    Samuele Salti$^{2}$\\
    $^1$Verizon Connect, Italy  ~~ $^2$ University of Bologna, Italy ~~ $^3$ University of Florence, Italy\\
    \small{\url{https://henriquepm.github.io/RendBEV/}}
    }

\maketitle
\copyrightnotice

\begin{strip}
    \begin{minipage}{\textwidth}\centering
        \vspace{-50pt}
    \includegraphics{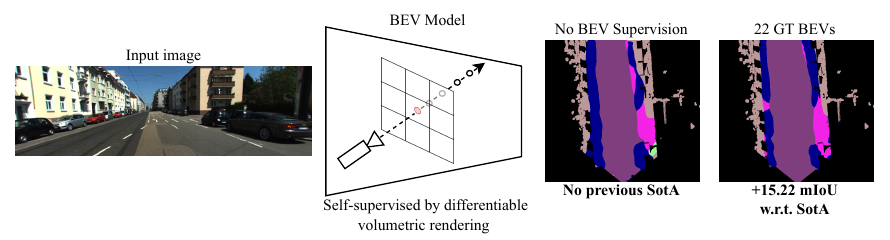}
    \captionsetup{type=figure}
    \vspace{-0.2cm}
    \captionof{figure}{\textit{\ourmethod} performs self-supervised training of any BEV semantic segmentation model via volumetric rendering. It enables to train  BEV semantic segmentation architectures in the absence of any labeled or pseudolabeled BEV data and provides state-of-the-art performance in the low-annotation regime when used  as a pretraining strategy.\label{fig:teaser}}
\vspace{-0.2cm}
\end{minipage}
\end{strip}
\begin{abstract}
    Bird's Eye View (BEV) semantic maps have recently garnered a lot of attention as a useful representation of the environment to tackle assisted and autonomous driving tasks. However, most of the existing work focuses on the fully supervised setting, training networks on large annotated datasets. In this work, we present \ourmethod, a new method for the self-supervised training of BEV semantic segmentation networks, leveraging differentiable volumetric rendering to receive supervision from semantic perspective views computed by a 2D semantic segmentation model. Our method enables zero-shot BEV semantic segmentation, and already delivers competitive results in this challenging setting. When used as pretraining to then fine-tune on labeled BEV ground truth, our method significantly boosts  performance in low-annotation regimes, and sets a new state of the art when fine-tuning on all available labels.
\end{abstract}
\section{Introduction}
\label{sec:intro}
The Bird's Eye View (BEV) is an orthographic projection of the world of great interest in assisted and autonomous driving and robotics. 
It has several useful properties: the size and shape of objects is not dependant on the viewpoint, the scale is consistent, distances are metric if the ground is flat and it is a suitable representation to fuse multiple sensors and modalities under a unified reference frame. The movement of road agents is mostly restricted to the ground plane, thus the BEV conveys most of the relevant information for road scene understanding while being more efficient than an explicit three-dimensional representation like a voxel grid. Many works in recent years address tasks in the Bird's Eye View such as semantic segmentation \cite{lu_monocular_2019, reiher_sim2real_2020, roddick_predicting_2020, philion_lift_2020, harley_simple-bev_2023, gosala_birds-eye-view_2022}, 3D object detection \cite{roddick_orthographic_2018, reading_categorical_2021, li_bevformer_2022} and trajectory prediction \cite{hu_fiery_2021}. Most of these works were enabled by the release of large scale multi-modal driving datasets like nuScenes\cite{caesar_nuscenes_2020}, KITTI-360\cite{liao_kitti-360_2023} or Waymo\cite{waymo_2020_CVPR} and are trained in a fully-supervised way. 
Obtaining ground truth BEV data is however a lengthy process which involves expensive sensors like lidar and heavy post-processing and labeling. In some circumstances, \eg{} when simple dashboard cameras (dashcams) are used to collect the video sequences, these sensors are not available during data acquisition, making it impossible to obtain these ground truth labels in a lot of economically relevant scenarios. Additionally, models trained on these datasets are not easy to transfer directly to other domains \cite{wang_towards_2023}. An alternative, recently proposed in \cite{gosala_skyeye_2023} is the usage of BEV pseudolabels. These pseudolabels present however certain disadvantages: they are cumbersome to compute, they require to bake in a lot of prior knowledge, and the pipeline to compute them has many moving parts and does not generalize across datasets. 

To make training of models for BEV semantic segmentation possible even in scenarios where annotation is too expensive or plainly unfeasible, we propose \textit{\ourmethod}, a method to self-supervise BEV semantic segmentation networks without using any BEV labels or pseudolabels. \ourmethod{} requires video sequences at training time and trains a monocular semantic segmentation BEV network by using its predictions for a frame to render semantic perspective views for some other video frames. A loss to update the BEV network can then be computed by getting pseudolabels in perspective view from an off-the-shelf 2D semantic segmentation model. 
Rendering of semantic segmentation in perspective view from the predicted BEV is made possible by leveraging recent advances in novel view synthesis \cite{mildenhall_nerf_2020}. In particular, we query density values from Behind the Scenes, a neural field pretrained in a self-supervised manner \cite{wimbauer_behind_2023}, and use it to perform differentiable rendering of class probabilities predicted by the BEV model.

The main contribution of this paper is the proposal of the first method to train Bird's Eye View semantic segmentation networks that uses no explicit BEV supervision (neither ground truth labels nor pseudolabels). The only data we assume available (aside from video sequences) are camera calibration and poses. We provide extensive experimental results on the KITTI-360 dataset. We show how our method can be used to train effectively recent BEV architectures in a completely self-supervised way, enabling them to deliver results already superior to some fully-supervised methods.    
We also present ablation studies to validate our design choices, showing the importance of rendering future frames in training sequences and the negligible impact of using frontal view pseudolabels instead of ground truth. 
Finally, we show that if BEV ground truth labels are available, our method can be used as pretraining, significantly boosting the performance of the segmentation network compared to training from scratch, especially in low-annotation regimes. With as low as 0.1\% -- a total of 22 examples -- of ground truth data available, our method provides results competitive with fully supervised models and significantly outperforms the state of the art in the low-annotation regime, as shown in \cref{fig:teaser}. 
When fine-tuning on 100\% of the trained data, pretraining with \ourmethod{} achieves state of the art results.

\section{Related Work}
\label{sec:related work}
\subsection{Bird's Eye View Segmentation from Images}

BEV segmentation pipelines from images generally follow multiple steps \cite{li_delving_2024}: first the perspective image(s) are passed through an encoder to obtain image features. Then, these features are lifted to a three-dimensional representation and collapsed or directly transformed to BEV. Finally, a network operating on the BEV features produces the final desired segmentation. Past work mainly focuses on the transformation between the frontal and the orthographic view. 
We can broadly categorize the existing models in four main groups according to the perspective to bird's-eye-view transformation method. 
\textbf{Neural Network-based}. Network-based models rely on a neural network to directly produce the BEV segmentation or transform image features into BEV features. VED \cite{lu_monocular_2019} uses a Variational Encoder-Decoder architecture to simultaneously solve the mapping and segmentation problems. VPN \cite{pan_cross-view_2020} employs a two-layer MLP to perform the view transformation. BEVFormer \cite{li_bevformer_2022} combines geometry-awareness with more sophisticated neural mechanisms like spatio-temporal deformable attention.
\textbf{Depth-based}. These approaches predict depth or depth distributions, either with a pretrained model \cite{reading_categorical_2021} or implicitly in the network architecture \cite{philion_lift_2020} to inform the model on the geometry of the scene and place features in an appropriate 3D location. 
\textbf{Homography-based}. These methods rely on warping the image or image features to the ground plane using an homography. A common baseline approach in BEV semantic segmentation is to to generate a perspective view semantic segmentation of the image and then warp it using the inverse perspective mapping (IPM) \cite{mallot_inverse_1991}. This process results in extreme deformation of areas which are not flat in the original perspective image. In some works like PanopticBEV \cite{gosala_birds-eye-view_2022} or Cam2BEV \cite{reiher_sim2real_2020} the IPM is used to warp the features or part of the features in the perspective view to bird's eye view transformation step.
\textbf{Parameter-free}. Simple-BEV \cite{harley_simple-bev_2023} proposes a parameter-free lifting mechanism, by projecting the centre points of a voxel grid to the perspective feature space and performing bilinear sampling, obtaining competitive results while using a simpler architecture compared to other methods.

These works rely on ground truth BEV semantic segmentation to provide supervision. However, obtaining these labels is an expensive process. It requires point clouds registered with the perspective images that have to be manually annotated. Recently, Gosala \etal \cite{gosala_skyeye_2023} proposed a training methodology that does not use BEV GT, comprised of two stages. First, the architecture without the BEV segmentation head is pretrained by \emph{predicting} future frontal view semantic labels. They are computed by coarsely approximating perspective projection, \ie by collapsing along the depth dimension a feature volume that the architecture is assumed to compute. Then, to provide supervision to train the BEV segmentation head and fine-tune the network, they need to introduce an explicit supervision step with BEV pseudolabels, and feed the head with features from the same volume, but collapsed vertically, to approximate the BEV orthographic projection. 

Hence, the head works on a view of the feature volume that has not been supervised while pretraining, making it less effective, especially in the low-annotation regime. In our work, we propose a new architecture-agnostic self-supervised training scheme based on the proper \emph{rendering} of future segmentation frames which does not require an explicit supervision step and thus can train the full BEV segmentation network without any BEV supervision (neither labels nor pseudolabels) and does not introduce approximations of the image formation process.
OccFeat \cite{sirko-galouchenko_occfeat_2024} authors propose a pretraining methodology for BEV segmentation networks which brings significant improvements in low-annotation regimes, but relies on access to lidar data for its occupancy-guided loss. Differently from OccFeat, in this work we focus on a setting with no lidar data.

\begin{figure*}[t]
     \centering
     \includegraphics[trim={0 0 2cm 0},clip,width=0.90\linewidth]{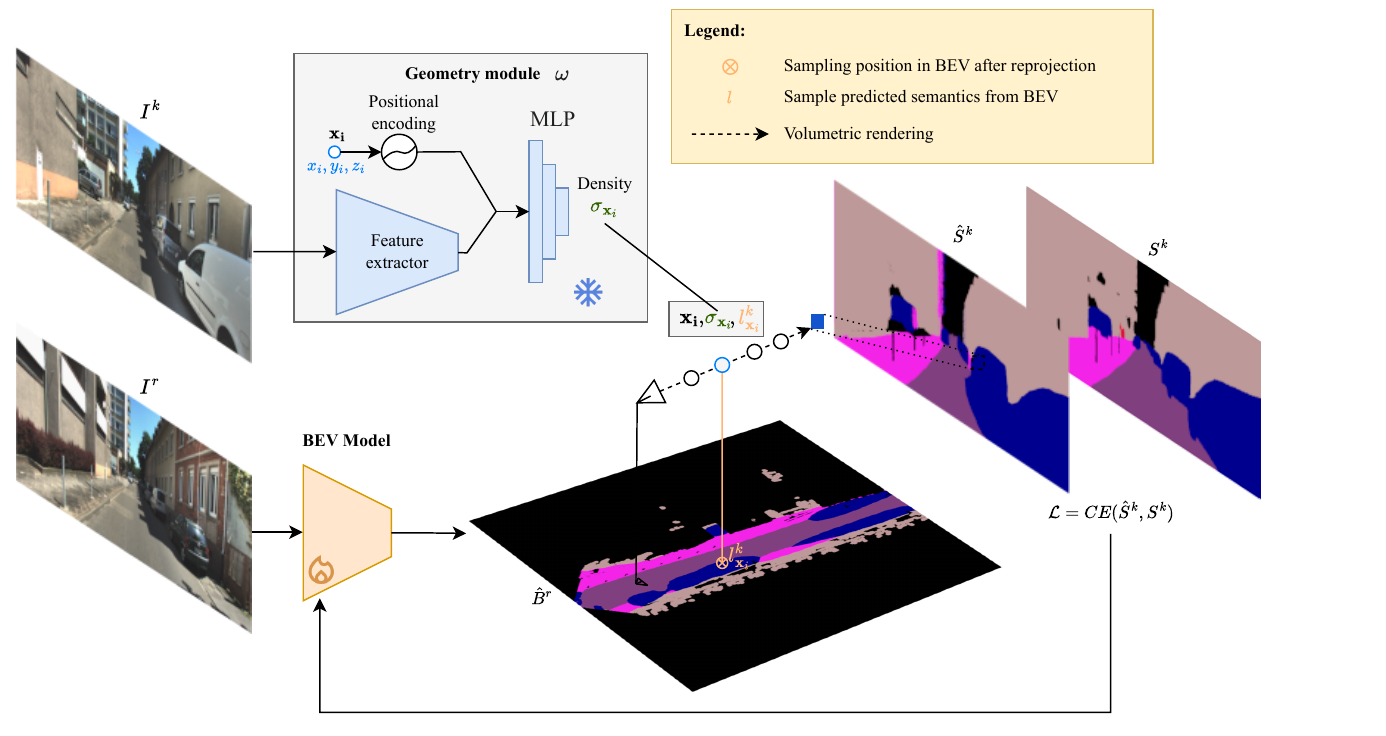}
     \caption{\ourmethod, our method for self-supervised training of BEV semantic segmentation models: we perform a forward pass with a reference view $I^r$ as input of the BEV network. We render the semantic semantic segmentation of \textbf{another} view $\hat{S}^k$, with class probability values $l^k_{\mathbf{x_i}}$ sampled from the BEV prediction $\hat{B}^r$ and densities $\sigma_{\mathbf{x}_i}$ queried from a pretrained frozen model $\omega$ that receives the target frame $I^k$ as input. We supervise the network with a cross entropy loss computed with the rendered semantic segmentation $\hat{S}^k$ and the target semantic segmentation $S^k$.}
     \label{fig:main_figure_method}
\end{figure*}

\subsection{Scene Reconstruction and Neural Fields}
The prediction of scene geometry from monocular cameras has been a field of great interest and progress in recent years. Self-supervised depth-from-mono models \cite{godard_digging_2019} learn to predict the depth from each pixel in a self-supervised way by exploiting geometric consistency in image sequences. These methods cast the problem as novel view synthesis, predicting the appearance of a target image from the viewpoint of another image and minimizing a photometric loss.

Inspired by Neural Radiance Fields \cite{mildenhall_nerf_2020}, Behind the Scenes \cite{wimbauer_behind_2023} proposes a method to obtain a more complete geometric representation of a scene than a per-pixel depth: an implicit density field. This field maps every point in the frustum of the input image to a density value.
The architecture is composed of an image encoder-decoder and a MLP, which are jointly optimized by an image reconstruction loss. The reconstruction is obtained with a differentiable volume rendering formulation. S4C \cite{hayler_s4c_2023} extends \cite{wimbauer_behind_2023} with an additional MLP that predicts class logits to tackle the task of semantic scene completion, with the addition of semantic segmentation pseudolabels as reconstruction targets. In this work, we leverage Behind the Scenes to perform differentiable volumetric rendering and create supervision, but we do not change its architecture to solve an additional task. In contrast, we use it to render class probabilities predicted by existing specialized BEV networks to make it possible to train them in a self-supervised way.
\section{The \textit{\ourmethod} method}
\label{sec:method}

Here we present \textit{\ourmethod}, a new architecture-agnostic method to train BEV segmentation networks in a self-supervised manner -- without using any BEV ground truth labels or pseudolabels. Our method, summarized in \cref{fig:main_figure_method}, relies on the key idea of sampling the output of a BEV semantic segmentation network to render the perspective view semantic segmentation image from a different timestamp. To make it possible, density values are sampled from a frozen model which predicts a density field. This model is trained in a previous step in a self-supervised way. We operate under the common assumption for self-supervised depth-from-mono models that the camera is moving and the scene is static.

At training time, we follow a setting based on Behind The Scenes \cite{wimbauer_behind_2023}. We have access to a sequence of  $N = \{1, 2, \dots, n\}$ monocular frontal view RGB frames $I^k, k\in N$, their corresponding semantic segmentations $S^k$ and camera poses with respect to an arbitrary world reference frame $M^{k\rightarrow w}$. 

Given a random frame in the sequence, $I^r$, we run the BEV network we wish to train on it, to generate class probabilities for each class, \ie{} the output of the final softmax layer, in each pixel of the BEV, $\hat{B}^r$. To supervise the network, we then consider another frame $I^k$ and
 reconstruct $P=\{1, 2, \dots, p\}$ patches from the semantic segmentation frame $S^k$ by performing volumetric rendering \emph{of class probabilities}.  
 To this end, we emit rays from every pixel in the patch and we sample $\textit{m}$ points $\textbf{x}_i, i=1,\dots,m$ along the ray (uniformly in disparity with an added random noise factor) to discretize the integral of volumetric rendering \cite{volumetric_rendering_Kajiya_1984}. Volumetric rendering needs a density value at each 3D point in space. We obtain it by querying the neural field described in Behind the Scenes, pretrained in a self-supervised way \cite{wimbauer_behind_2023}.

Hence, for each point $\textbf{x}_i$ in a ray going through pixel $(u,v)$, we query the volumetric density value $\sigma_{\mathbf{x}_i}$ from a frozen model $\omega$. Differently from \cite{wimbauer_behind_2023}, we compute features in $\omega$ from the frame from which the ray is cast, which is shown to be key 
 for our use case in the ablation studies.
 In particular, let $\delta_i$ be the distance between $\textbf{x}_i$ and $\textbf{x}_{i+1}$, and $\alpha_i$ be the probability of a ray hitting a surface in a 3D position between $\textbf{x}_i$ and $\textbf{x}_{i+1}$, \ie{}
 \begin{equation}
    \alpha_i = 1 - \exp( \sigma_{\textbf{x}_i} \delta_i) \mbox{.}
\end{equation}
 Given the previous $\alpha_j, j=1,\dots,i-1$, along a ray, we can compute the probability $T_i$ that the ray travels in free space before $\textbf{x}_i$ as

\begin{equation}
    T_i = \prod_{j=1}^{i-1} (1 - \alpha_j) \mbox{.}%
\end{equation}

This is routinely used in novel view synthesis to decide the color $\hat{c}$ of a pixel by integrating colors of the 3D points $c_{\textbf{x}_i}$ along the ray as
\begin{equation}
    \hat{c} = \sum_{i=1}^{m}T_i \alpha_i c_{\textbf{x}_i}
\end{equation}

Since our aim is to render class probability, we need to associate a vector of class probabilities to each point in 3D space. These values should come from the predicted BEV for $I^r$ so that it can be supervised by the rendering. Thus, we 
sample class probability distribution values from the network-generated BEV semantic segmentation $\hat{B}^r$ of the reference image $I^r$. We do so by transforming the 3D points $\mathbf{x}_i$ to its 3D frame using camera poses $M^{k\rightarrow r}={(M^{r\rightarrow w})}^{-1} M^{k\rightarrow w}$ and orthographically projecting the transformed points $\mathbf{x}_i$ to the BEV, \ie{} dropping the vertical coordinate $y$, with the projection (in homogeneous coordinates):
\begin{equation}
    \pi_{\perp} 
 = \begin{bmatrix}
1 & 0 & 0 & 0\\
0 & 0 & 1 & 0\\
0 & 0 & 0 & 1\\
\end{bmatrix} 
\end{equation}

Therefore, we obtain a class probability $l_{\mathbf{x}_i}^k$ for each point $\textbf{x}_i$ along a ray cast from frame $k$, by the operation described in \cref{eq:semantics_sampling}:
\begin{equation}
    \label{eq:semantics_sampling}
    l_{\mathbf{x}_i}^{k} =\hat{B}^r \langle \pi_{\perp} ( M^{k\rightarrow r} \textbf{x}_i ) \rangle
\end{equation}
where $\langle \cdot \rangle$ is the nearest neighbor sampling operator.   
This scheme relies on the assumption that the class is constant across the pillar stemming from each position in the BEV. There are cases where this assumption does not hold \eg when part of an object ``floats'' above another, like a building's balcony above a sidewalk or a tree canopy extending above a road. However, we consider it a good enough approximation for our application.
We follow \cite{siddiqui_panoptic_2022} in applying the softmax prior to rendering and not afterwards. Otherwise, the unbounded nature of the logits produced by the network could lead to violations of the geometric constraints imposed by the neural density field.

Finally, we obtain the class probability prediction for the pixels $(u,v)$ in a patch in the target frame $k$ using the rendering equation with the previously computed probabilities for each 3D point along the ray:
\begin{equation}
    \hat{l}^{k}_{u,v} = \sum_{i=1}^{m}T_i \alpha_i l_{\textbf{x}_i}^{k}
\end{equation}

Our loss function is then a class-weighted cross-entropy between the prediction and the semantic segmentation label in $S_k$ at pixel $(u,v)$ aggregated across the sampled patches. 

\begin{equation}
    \mathcal{L}^{k}_{u,v} =  WCE(\hat{l}_{u,v}^{k}, S^{k}_{u,v})
\end{equation}

The total loss for a frame is the average of the losses for all pixels of all patches. 

Points along the rays we cast might fall outside of the area where the BEV semantic segmentation of the reference image $\hat{B}^{r}$ is defined, thus not having valid values to sample. Including rays with a lots of these points in the supervision could negatively affect the training, thus we also perform volumtric rendering of an indicator variable for the 3D point falling outside the reference BEV $\hat{B}^r$ and we filter out rays for which the rendered value exceeds a certain threshold $\tau$. 

\begin{figure}
    \centering
    \includegraphics[trim={0 5.2cm 0 0},clip,width=\linewidth]{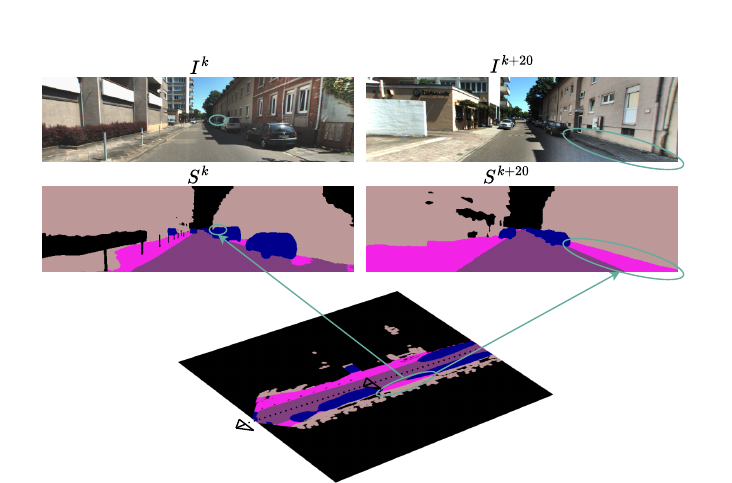}
    \caption{Importance of rendering temporally far frames with respect to the reference one: by reconstructing future frames we supervise areas which are occluded in the reference frame and provide denser supervision in spatially faraway areas which otherwise would only be supervised by a very small amount of pixels in the reference frame.}
    \label{fig:secondary_figure_method}
\end{figure}

Given a sequence, we use multiple frames to supervise the BEV at a reference $I^r$ and we average the loss across them. It is common practice, \eg{} in the self-supervised depth-from-mono literature \cite{godard_digging_2019}, to use as frames to compute self-supervised losses adjacent frames in a video sequence, \ie{} let $k$ be either $r-1$ or $r+1$. However, letting $k$ vary only in this close range would be detrimental in our scenario. As shown in \cref{fig:secondary_figure_method}, to be able to learn how to segment regions blocked or faraway in the reference frame, we need to use temporally far future frames. 

\section{Experiments}
\label{sec:experiments}
\subsection{Dataset}
We perform our experiments on the large scale KITTI-360 dataset \cite{liao_kitti-360_2023}. We use as BEV ground truth (for evaluation, fine-tuning, and baselines) the labels provided by the authors of SkyEye \cite{gosala_skyeye_2023}. We follow their training-validation split, both on the full datset and to measure performance at different data regimes.  We perform zero-shot generalization experiments on the Waymo-BEV dataset \cite{waymo_2020_CVPR, gosala_skyeye_2023}, a more challenging dataset with respect to KITTI-360 with higher variance in terms of geographic, climatic and lighting conditions.

\subsection{Experimental setting}
Our method is architecture agnostic. For the experiments presented in this work we use as geometry module a Behind the Scenes model~\cite{wimbauer_behind_2023} trained on the KITTI-360 dataset\cite{liao_kitti-360_2023} in a self-supervised way. We select the architecture described in SkyEye \cite{gosala_skyeye_2023} as BEV semantic segmentation network. In the supplementary material, we provide further details on our experimental details, an overview of SkyEye's architecture and method as well as additional results using Simple-BEV \cite{harley_simple-bev_2023} instead.

To the best of our knowledge, no other method is capable of providing results without using any BEV label or pseudolabel. We construct an unsupervised baseline approach via an IPM\cite{mallot_inverse_1991}, which we compute using the known camera height and extrinsic parameters, assuming the ground is a flat and horizontal plane. The computed warping is applied to semantic segmentation images to generate the target BEV. All perspective view semantic segmentation images -- both for the IPM and our pretraining -- are generated by an off-the-shelf Panoptic-Deeplab \cite{cheng2020panoptic} model pretrained on the Cityscapes \cite{cordts_cityscapes_2016} dataset and were made available by the authors of S4C \cite{hayler_s4c_2023}.

\subsection{Quantitative results}
\begin{table*}
\centering
\caption{Evaluation of BEV semantic segmentation on the KITTI-360 dataset. All results but \ourmethod, Simple-BEV and IPM (run by us on the same splits) from \cite{gosala_skyeye_2023}.}
\vspace{-0.2cm}
\label{tab:main_results}
\scriptsize
 \setlength\tabcolsep{3.7pt}
 \begin{tabular}{c|c|cccccccc|c}
 \toprule
 \textbf{BEV (\%)} & \textbf{Method} & \textbf{Road} & \textbf{Sidewalk} & \textbf{Building} & \textbf{Terrain} & \textbf{Person} & \textbf{2-Wheeler} & \textbf{Car} & \textbf{Truck} & \textbf{mIoU} \\
 \midrule
 \multirow{2}{*}{0} & IPM \cite{mallot_inverse_1991} &58.39 & 23.07 & 12.55 & 32.47 & 0.58 & \textbf{1.44} & 11.61 & \textbf{5.16} &  18.16 \\
  & \ourmethod (ours) & \textbf{68.34} & \textbf{33.27}  & \textbf{33.26} & \textbf{44.60} & \textbf{1.23} & 0.72 & \textbf{32.37} & 3.39 & \textbf{27.15}\\
  \midrule

   \multirow{2}{*}{1} &SkyEye \cite{gosala_skyeye_2023} & 70.69 & 31.13 & 32.38 & 40.08 & 0.00 & 0.00 & 29.08 & 3.95 & 25.91 \\
   & \ourmethod (ours) &  \textbf{74.76} & \textbf{40.20} & \textbf{41.07}  & \textbf{46.40} & \textbf{1.67} & \textbf{3.94} & \textbf{35.78} & \textbf{5.90} & \textbf{31.22}  \\
  \midrule
  \multirow{7}{*}{100}& TIIM~\cite{saha_translating_2022}  &63.08 & 28.66 & 13.70 & 25.94 & 0.56 & 6.45 & 33.31 & 8.52 & 22.53 \\
  & VED~\cite{lu_monocular_2019} & 65.97 & 35.41 & 37.28 & 34.34 & 0.13 & 0.07 & 23.83 & 8.89 & 25.74 \\
  & VPN~\cite{pan_cross-view_2020} & 69.90 & 34.31 & 33.65 & 40.17 & 0.56 & 2.26 & 27.76 & 6.10 & 26.84 \\
  & PON~\cite{roddick_predicting_2020} &  67.98 & 31.13 & 29.81 & 34.28 & 2.28 & 2.16 & 37.99 & 8.10 & 26.72 \\
    & Simple-BEV~\cite{harley_simple-bev_2023} & 70.66 & 35.50 & 34.67 & 41.18 & 1.04 & 2.11 & 38.24 & 12.42 & 29.48 \\
  & PoBEV~\cite{gosala_birds-eye-view_2022} & 70.14 & 35.23 & 34.68 & 40.72 & 2.85 & 5.63 & 39.77 & 14.38 & 30.42 \\
  & SkyEye \cite{gosala_skyeye_2023}  &72.82 & 38.27 & 40.86 & \textbf{45.86} & \textbf{3.59} & \textbf{7.74} & 41.37 & 9.74 & 32.53 \\
 & \ourmethod (ours)  & \textbf{74.83} & \textbf{40.98} & \textbf{41.80} & 45.63 & 3.47 & 6.09 & \textbf{45.55} & \textbf{16.74}  & \textbf{34.39} \\
 \bottomrule
 \end{tabular}

\end{table*}
In \cref{tab:main_results} we present the main results of our evaluation.  We report results under three scenarios: using no BEV annotations; when a small quantity (1\%) of GT labels is available; when all training data are annotated. In the last case, we include in the evaluation several supervised baselines: TIIM\cite{saha_translating_2022}, VED\cite{lu_monocular_2019}, VPN\cite{pan_cross-view_2020}, PON\cite{roddick_predicting_2020}, a modification of Simple-BEV\cite{harley_simple-bev_2023} to perform general semantic segmentation instead of vehicle segmentation.

Without explicit BEV supervision (either labels or pseudolabels), only \ourmethod{} and the unsupervised IPM baseline can produce a BEV semantic map. The model trained with our methodology at the 0\% BEV regime outperforms the baseline and already delivers very good results, even better than four of the fully-supervised models.  The network is capable of producing good quality segmentations for the static classes. In the dynamic classes (person, 2-wheeler, car and truck), our model has a wider performance gap with fully supervised models, which is reasonable since it relies on a static scene assumption. Yet, it already provides acceptable performance for the car class.

When a small quantity of GT labels is available, the state of the art is SkyEye \cite{gosala_skyeye_2023}. To perform a fair comparison, we consider its results when pretraining is performed by using perspective view pseudolabels, as done by \ourmethod{}. \ourmethod{} outperforms SkyEye by a significant margin. Additionally, it delivers overall very good performance, better than  strong fully supervised models trained on 100 times more data such as Simple-BEV and PoBEV. This result demonstrates its effectiveness in enhancing model performance in low-annotation regimes, probably due to the fact that our training methodology provides an already solid starting point to the full BEV segmentation network, as -- differently from the two-stage procedure in SkyEye -- it is capable of training the BEV semantic segmentation head prior to any BEV explicit supervision. When fine-tuning on all the training data, we compare again against SkyEye and the fully supervised baselines. Pretraining with \ourmethod{} is the best approach.

In \cref{tab:pretraining-preds-finuting-percentages} we present a more detailed study on the performance of our methodology when used as pretraining followed by fine-tuning at different annotation regimes. We compare the results obtained with the same architecture -- in all cases the one proposed in \cite{gosala_skyeye_2023} -- when trained from scratch, when SkyEye's pretraining is applied, and when \ourmethod{} is used as pretraining. Our method achieves competitive results fine-tuning with as few as \(\sim \)0.1\% of the training data (a total of 22 images) and produces better results across the board than both SkyEye's pretraining strategy and the same model trained from scratch. As expected, the gap becomes narrower as the total amount of GT data available becomes larger and the overall importance of the pretraining in the whole process decreases.

\begin{table*}
\centering
\caption{Impact of pretraining at different annotation regimes. All scores are reported on the KITTI-360 dataset.}
\vspace{-0.2cm}
\label{tab:pretraining-preds-finuting-percentages}
\scriptsize
\setlength\tabcolsep{3.7pt}
 \begin{tabular}{c|c|cccccccc|c}
 \toprule
 \textbf{BEV~(\%)} & \textbf{Pretraining} & \textbf{Road} & \textbf{Sidewalk} & \textbf{Building} & \textbf{Terrain} & \textbf{Person} & \textbf{2-Wheeler} & \textbf{Car} & \textbf{Truck} & \textbf{mIoU} \\
 
 \midrule
\multirow{3}{*}{0.1}& -- & 56.43 & 19.95 & 23.64 & 7.17 & 0.00 & 0.00 & 12.59 & 0.00 & 14.97 \\
 & SkyEye & 57.35 & 19.36 & 22.13 & 12.54 & 0.00 & 0.00 & 10.56 & 0.00 &  15.24 \\
& \ourmethod & \textbf{74.81} & \textbf{40.10} & \textbf{39.19} & \textbf{45.67} & \textbf{1.50} & \textbf{1.67} & \textbf{34.76} & \textbf{6.00} & \textbf{30.46} \\

 \midrule
\multirow{3}{*}{1} & -- & 61.01 & 22.68 & 27.81 & 23.69 & 0.00 & 0.00 & 31.31 & \textbf{6.32} & 21.60 \\
 & SkyEye & 70.69 & 31.13 & 32.38 & 40.08 & 0.00 & 0.00 & 29.08 & 3.95 & 25.91 \\
& \ourmethod & \textbf{74.76} & \textbf{40.20} & \textbf{41.07}  & \textbf{46.40} & \textbf{1.67} & \textbf{3.94} & \textbf{35.78} & 5.90 & \textbf{31.22}  \\

 \midrule
 \multirow{3}{*}{10} & -- & 73.39 & 37.49 & 35.87 & 40.30 & \textbf{4.72} & \textbf{7.44}  & \textbf{44.64} & \textbf{12.23} & 32.01 \\ 
& SkyEye & 73.16 & 37.08 & 38.41 & 45.45 & 3.66 & 6.69 & 40.60 & 7.94 & 31.62 \\
& \ourmethod & \textbf{75.88} & \textbf{41.35} & \textbf{42.00} & \textbf{45.91} & 2.92 & 5.68 & 44.06 & 11.54 & \textbf{33.67} \\ 

\midrule
\multirow{3}{*}{50} & -- & 75.30 & 40.61 & 41.79 & 45.34 & 2.88 & 6.64 & \textbf{45.52} & 13.46 & 33.94  \\ 
 & SkyEye & 72.50 & 36.92 & 39.41 & 45.12 & \textbf{3.63} & \textbf{7.46} & 41.21 & 9.73 & 32.00 \\
& \ourmethod & \textbf{75.53} & \textbf{40.96} & \textbf{43.73} & \textbf{46.52} & 3.27 & 5.60 & 45.16 & \textbf{14.30} & \textbf{34.38}  \\ 
\midrule
\multirow{3}{*}{100} & -- & 73.01 & 37.78 & 39.15 & 43.68 & \textbf{5.44} & \textbf{10.76} & 45.41 & 12.25 &  33.72 \\
& SkyEye & 72.82 & 38.27 & 40.86 & \textbf{45.86} & 3.59 & 7.74 & 41.37 & 9.74 & 32.53 \\
 & \ourmethod & \textbf{74.83} & \textbf{40.98} & \textbf{41.80} & 45.63 & 3.47 & 6.09 & \textbf{45.55} & \textbf{16.74}  & \textbf{34.39} \\
 \bottomrule
 \end{tabular}

\end{table*}

To assess the generalization capabilities of  our method, we run a zero-shot evaluation experiment on the Waymo-BEV dataset \cite{waymo_2020_CVPR,gosala_skyeye_2023} of models trained on the KITTI-360 dataset. In \cref{tab:zero-shot} we present the obtained results. Even if we observe a drop in performance with respect to the one obtained on KITTI-360, the network is still able to produce good results, especially for road layouts and cars. 

\begin{table}[]
    \scriptsize
    \setlength\tabcolsep{2.7pt}
    \centering
    \caption{Zero-shot evaluation of \ourmethod{} on the Waymo-BEV dataset. Models trained on 0\% and 100\% GT labels from KITTI-360.  We also include results on KITTI-360 for reference (subset of classes available in Waymo-BEV).}
    \vspace{-0.2cm}
    \begin{tabular}{c|c|cccccc|c}
    \toprule
    \textbf{Val} & \textbf{BEV(\%)}&\textbf{Road} & \textbf{Sidew.} & \textbf{Build.}  & \textbf{Pers.} & \textbf{2-Wh.} & \textbf{Car} & \textbf{mIoU} \\
    \midrule
\multirow{2}{*}{Waymo-BEV}&    0&67.33 &9.91&21.24&3.08&0.17&22.86&20.76\\
& 100 &73.64&14.97&22.10&5.19&0.64&31.41&24.67\\
    \midrule
\multirow{2}{*}{KITTI-360}& 0 & 68.34 & 33.27  & 33.26 & 1.23 & 0.72 & 32.37 & 28.20 \\
& 100 & {74.83} & {40.98} & {41.80} & 3.47 & 6.09 & {45.55} &  {35.45} \\
    \bottomrule
    \end{tabular}
    \label{tab:zero-shot}
\end{table}

\subsection{Qualitative results}
\begin{figure*}[htp]
   \centering
   \includegraphics[width=0.85\linewidth]{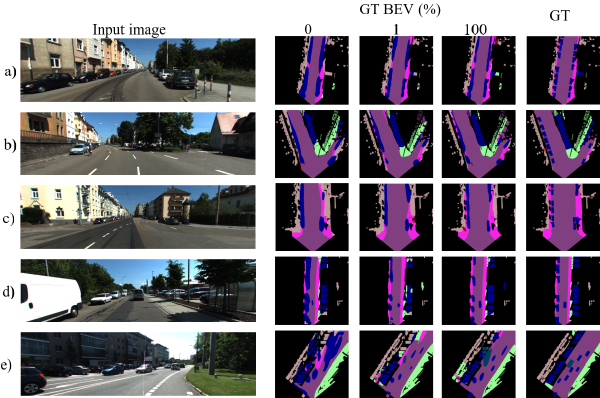}
   \caption{Qualitative results of our model at different annotation regimes and GT BEV semantic segmentation}
   \label{fig:qualitative_results_comparison}
\end{figure*}

We present qualitative results to illustrate the behavior of our model in different situations. In \cref{fig:qualitative_results_comparison} we provide a comparison of the qualitative results provided by our model when trained in a fully self-supervised fashion (0\% BEV) and when fine-tuned on 1\% and 100\% of the training data with the ground truth BEV semantic segmentation. We use the Cityscapes color palette, and mask out pixels not taken into account during evaluation (outside of the field of view, unlabeled or labeled with an irrelevant class). We observe that the model trained with no GT BEV is able to segment the scene layout in an already reasonably accurate way as well as to position objects, especially in more static scenes like a), b) and c) in \cref{fig:qualitative_results_comparison}. However, it struggles to differentiate the precise shape of cars (in blue) even when they are more clearly separated as in d) and fails in doing so when presented with more complex, dynamic scenes as in e). The model fine-tuned on 1\% of GT BEV provides better results, especially in terms of the shape of objects and in more dynamic scenes. While the model is not able to separate rows of  parked cars into individual masks (first three examples), the edge of the area labeled as car becomes sharper and less blob-like. In the scenes with moving vehicles d) and e) there is a great improvement in  segmentation quality, further accentuated when fine-tuning with 100\% GT.

\subsection{Ablations}
\begin{figure}[bp]
    \centering  \includegraphics[width=0.85\linewidth]{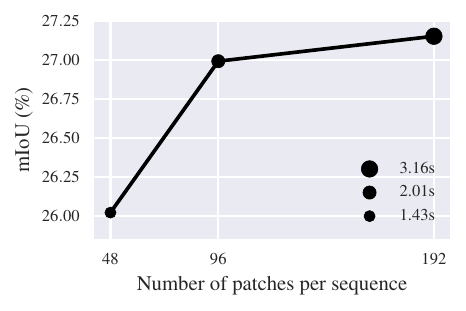}
    \caption{Effect of the number of patches per sequence on the mIoU. The marker size is proportional to the time per iteration during training.}
    \label{fig:npatches-miou}
\end{figure}
We investigate the effect of different choices in our pipeline by performing additional training experiments. 
In \cref{fig:npatches-miou} we show the effect on the mIoU of changing the number of patches sampled per sequence. A higher number of patches results in denser supervision, which translates to better performance in the ranges we study, but also to an increased time per training iteration. 

\begin{table*}
\centering
\caption{Impact of sampling patches from future frames and of feeding them to the geometry module}
\vspace{-0.2cm}
\label{tab:ablation_future_frames}
\scriptsize
\setlength\tabcolsep{3.7pt}
 \begin{tabular}{cc|cccccccc|c}
 \toprule
 \textbf{Future frames} & \textbf{$\omega$ on future frames} & \textbf{Road} & \textbf{Sidewalk} & \textbf{Building} & \textbf{Terrain} & \textbf{Person} & \textbf{2-Wheeler} & \textbf{Car} & \textbf{Truck} & \textbf{mIoU} \\
 \midrule
 \xmark & --& 49.52 & 25.45 & 26.68 & 43.95 & 1.19 & 1.35 & 19.65 & 8.03 & 21.98  \\
  \cmark & \xmark & 55.34 & 29.56 & 26.03 & 40.94 & 1.36 & 2.40 & 17.94 & 5.21 & 22.35   \\
 \cmark & \cmark & 65.29 & 31.54 & 28.58 & 43.46 & 2.15 & 2.67 & 28.31 & 6.17 &   26.02 \\
 \bottomrule
 \end{tabular}
 \vspace{-0.2cm}
\end{table*}

\begin{figure*}[htp]
    \centering
    \includegraphics[width=0.8\linewidth]{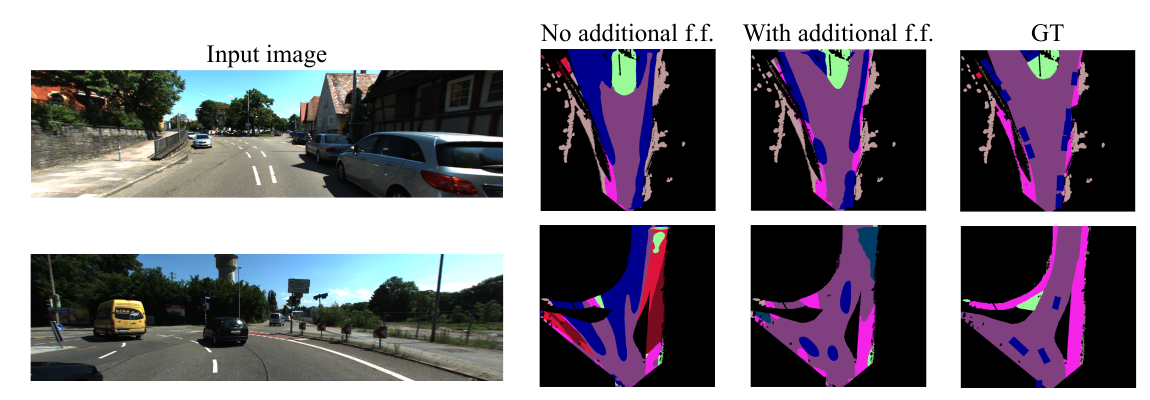}
   \caption{Comparison of the qualitative results of models trained with or without additional future frames.}
   \label{fig:qualitative_future_vs_nofuture}
\end{figure*}

To validate the importance of receiving supervision from future frames, we perform an experiment removing them, \ie{} we sample patches from timestamps $T= \{r-1, r+1\}$. We run this experiment sampling 48 patches per sequence. We present the results in \cref{tab:ablation_future_frames}, where the third row is the full \ourmethod{} method. 
Results on the first row show how removing future frames severely degrades performance, likely because the BEV semantic segmentation model lacks proper supervision in far areas in the BEV and in occluded areas. 
We also validate the importance of running the geometry module $\omega$ on the frame to be rendered as opposed to the reference frame used in Behind the Scenes. 
Adding future frames while still computing features from the reference frame (second row) gives worse results than the full \ourmethod{} strategy. Indeed, it does not make full use of the frames since the density predictions become less accurate as we query the network at 3D points more distant from the reference frame, and occupancy shadows cast by solid objects contaminate the prediction in areas occluded in the reference frames. Qualitative evidence is reported in \cref{fig:qualitative_future_vs_nofuture}. The model trained without additional future frames provides a reasonable segmentation in the region closer to the camera, but it is not able to predict the logic direction of the road or resolve occlusions even in situations that do not appear challenging. Due to the static scene assumption, it could be expected that the model trained without future frames fares better at least in scenes with dynamic objects, such as the second one in \cref{fig:qualitative_future_vs_nofuture}; however, the model suffers greatly from occupancy shadows and hallucinations in occluded areas, lacking the proper supervision to handle them.

To study the effect of using frontal view semantic segmentation ground truth, instead of the pseudolabels from an off-the-shelf model we employ in all other experiments, we train our model using the 2D semantic segmentation masks available in KITTI-360. We present the results in \cref{tab:ablation_FV_semseg}. We observe a negligible benefit in the performance when using ground truth, showing that our method is quite robust to the quality of the frontal view masks. 

\begin{table*}[tp]
\centering
\caption{Impact of using GT or model output in the frontal view semantic segmentation}
\vspace{-0.2cm}
\label{tab:ablation_FV_semseg}
\scriptsize
\setlength\tabcolsep{3.7pt}
 \begin{tabular}{c|cccccccc|c}
 \toprule
  \textbf{GT FV} & \textbf{Road} & \textbf{Sidewalk} & \textbf{Building} & \textbf{Terrain} & \textbf{Person} & \textbf{2-Wheeler} & \textbf{Car} & \textbf{Truck} & \textbf{mIoU} \\
 \midrule
 \cmark & 66.54 & 33.14 & 31.32 & 45.71 & 2.11 & 3.40 & 30.05 & 6.30 &  27.32 \\ 
 \xmark & 68.34 & 33.27  & 33.26 & 44.60 & 1.23 & 0.72 & 32.37 & 3.39 & 27.15  \\
 \bottomrule
 \end{tabular}

\end{table*}

\section{Conclusion}
\label{sec:conclusion}
In this paper we introduced \ourmethod, a novel method to train Bird's Eye View semantic segmentation networks in a self-supervised manner. To this end, we leveraged a pretrained neural density field and an off-the-shelf semantic segmentation model working in perspective view. Experimental results on KITTI-360 validate the competitiveness of the method, both as a stand-alone self-supervised training technique with no BEV supervision, and as a pretraining stage when some amount of ground truth data is available. With as little as 0.1\% of BEV annotations, \ourmethod{} already delivers competitive performance, and it provides state of the art results when 100\% annotations are used.

Future research should focus on the explicit modeling of dynamic objects to enhance performance in more complicated scenes. This could be achieved by \eg integrating trajectory estimation methods in the training framework. The extension of the method to multi-camera rigs is another interesting avenue.

\onecolumn
\twocolumn
{\small
\bibliographystyle{ieee_fullname}
\bibliography{ss_bev}
}
\onecolumn
\twocolumn
\input{sup}

\end{document}

%% file: sup.tex

\pagebreak

\setcounter{page}{1}
\setcounter{table}{0}
\setcounter{figure}{0}
\makeatletter

\renewcommand{\thetable}{S\arabic{table}}
\renewcommand{\thefigure}{S\arabic{figure}}
\setcounter{section}{0}
\renewcommand{\thesection}{S-\arabic{section}}

\begin{strip}
    \begin{center}
        \textbf{\Large\ourmethod: Semantic Novel View Synthesis for \\ Self-Supervised Bird's Eye  View  Segmentation}
        
      \vspace{5ex}
    \end{center}
    \begin{center}
        \textbf{\large-- Supplementary Material --}
    \end{center}
\end{strip}

In this supplementary material, we present additional explanations, experiments and results to complement the main paper. 

\section{SkyEye's Methodology}

In this section we present a brief overview of SkyEye's work and its key differences with our proposal. We would like to distinguish between their proposed network architecture (which we use in our experiments) and their proposed training framework (which we compare against).
For a more detailed report, we refer the interested reader to the original paper by Gosala et al. \cite{gosala_skyeye_2023}.

\subsection{Network Architecture}
SkyEye's model is composed of four main items. a) A 2D image encoder that extracts features from the input images. Chosen to be Efficient-D3's backbone in SkyEye's main experiments (and ours). b) A lifting module which populates a 3D voxel grid with features. c) A frontal view semantic head used in their implicit supervision. This module is not used in our experiments. d) A BEV semantic segmentation head. Additionally, an independent depth network is used to generate the the pseudolabels for their explicit supervision. This depth network is not used in our experiments.

\subsection{Training Framework}
SkyEye's proposed training framework comprises two stages: a pretraining referred to as implicit supervision and a final stage in which the actual BEV semantic segmentation head is trained: the explicit supervision. We give an overview of these two types of supervision and then contrast with our proposed training framework, highlighting differences and similarities.

\textbf{Implicit Supervision.} In the implicit supervision stage, the supervision signal provided by static elements in the scene is exploited by enforcing consistency. This is done by predicting the semantic segmentation of future timestamps in frontal view using only the 3D features computed from the initial frame. A cross-entropy loss is computed between target frontal view semantic segmentation labels and predicted values. A weight factor modulates the contribution of each frame from the sequence, linearly decaying from $1$ to $0.2$.

\textbf{Explicit Supervision.} During the implicit supervision stage, the BEV segmentation head is not trained. In order to circumvent the necessity of training the network with GT BEV annotations, SkyEye's authors propose a pseudolabel generation pipeline. This pipeline is based on an independent depth-from-mono network, a DBSCAN-based instance generation module and a densification module based on morphological operations. 

\textbf{Comparison with \ourmethod.} Similarly to SkyEye, \ourmethod\space also tres to exploit spatiotemporal consistency with a static scene assumption to train a BEV segmentation network. However, two main differences with our method exist, one conceptual and another methodological. At the conceptual level, our method can be run without the explicit supervision stage, avoiding the usage of BEV labels or pseudolabels. This is achieved with a methodological difference in the way of providing  ``self-supervision'' to the network. Instead of \emph{predicting} the semantic segmentation of future timesteps with features generated from the initial timestamp (which enables SkyEye's pretraining method to supervise part of their network but not their BEV semantic segmentation head), we \emph{render} the semantic segmentation of future timestemps, using class probability values sampled from the output of the BEV network. This lets gradient flow through the BEV semantic segmentation head already at this stage.

This difference grants the possibility to train models in a setting where no BEV supervision in any form is available and gives a good starting point for training if some GT BEV labels are available. We hypothesize that the performance gains (especially at low annotation regimes) when fine-tuning on GT BEV labels, are thanks to the capacity of our method to already provide supervision in the previous step to the semantic segmentation head and thus having a more advantageous starting point with respect to training it from scratch.

\section{Experiments with Simple-BEV}
\label{sec_sup:Simple-BEV_results}
We execute a supplementary set of experiments to validate \ourmethod\space with a different architecture for the BEV semantic segmentation network. To this end, we modify Simple-BEV \cite{harley_simple-bev_2023} and adapt it to our setting, by making it work with monocular frontal images only, increasing the number of classes in its semantic segmentation head and removing the auxiliary task heads. We train the network using the RendBEV method and then fine-tune the model at different percentage splits of the dataset. To provide a baseline comparison, we train the same model from scratch on the same splits. We present the results obtained in these experiments in \cref{tab_sup:Simple-BEV-gt}. The performance we reach while using \ourmethod\space as a standalone training is slightly inferior to the one obtained with SkyEye's architecture as BEV semantic segmentation network, but still competitive. When fine-tuning on available ground truth, \ourmethod\space proves to be useful as pretraining in the lower annotation regimes. When the amount of ground truth data is high (in the 50\% and 100\% splits) the pretrained models obtain overall performances almost equal to the ones trained from scratch in terms of mIoU. 

\begin{table*}
\centering
\caption{Study of the performance of our method with Simple-BEV as BEV semantic segmentation network at different annotated data regimes. All scores are reported in the KITTI-360 dataset.}
\label{tab_sup:Simple-BEV-gt}
\footnotesize
\setlength\tabcolsep{3.7pt}
 \begin{tabular}{c|c|cccccccc|c}
 \toprule
 \textbf{BEV~(\%)} & \textbf{Pretraining}  & \textbf{Road} & \textbf{Sidewalk} & \textbf{Building} & \textbf{Terrain} & \textbf{Person} & \textbf{2-Wheeler} & \textbf{Car} & \textbf{Truck} & \textbf{mIoU} \\
 \midrule
 \multirow{1}{*}{0.0} & RendBEV  & 65.46 & 30.30 &  29.49  & 38.46 & 1.94 & 2.49 & 30.92 & 7.17 & 25.78 \\
 \midrule
\multirow{2}{*}{0.1}  & -- & 45.78 & 14.01 & 11.35 & 4.22 & 0.12  & 0.25 & 5.87 & 4.60 & 10.26 \\
  & RendBEV & \textbf{67.19} & \textbf{32.60} & \textbf{32.39} & \textbf{39.11} & \textbf{1.92} & \textbf{2.69} &\textbf{32.30} & \textbf{7.60} & \textbf{26.98} \\
\midrule
\multirow{2}{*}{1} & -- & 57.45 & 23.16 & 19.34 & 21.37 & 0.06 & 0.11 &  18.20 & 1.52 &  17.65\\
 & RendBEV & \textbf{68.84} & \textbf{34.73} & \textbf{32.76} & \textbf{38.66} & \textbf{2.18} & \textbf{3.07} &  \textbf{34.27} & \textbf{5.18} &  \textbf{27.46}\\
 \midrule

\multirow{2}{*}{10} & -- & 70.42 & 34.37 & 30.28 & 35.36 & 0.3 &  0.84 & 34.43 & \textbf{10.03} & 27.00 \\
 & RendBEV & \textbf{70.66} & \textbf{36.13} & \textbf{36.34} & \textbf{40.02} & \textbf{1.66} & \textbf{4.91} & \textbf{35.80} & 5.74 &  \textbf{28.90}\\
\midrule
 
\multirow{2}{*}{50} & -- & \textbf{72.05} & 35.51 & 34.92 & 37.36 & 1.01 & 1.51 & \textbf{38.59} & \textbf{11.64} & 29.07 \\
 & RendBEV  & 70.70 & \textbf{36.00} & \textbf{36.73} & \textbf{40.38} & \textbf{1.72} & \textbf{5.17}  & 36.63 & 6.07 & \textbf{29.18} \\
\midrule
 \multirow{2}{*}{100} & --  & \textbf{70.66} & 35.50 & 34.67 & \textbf{41.18} & 1.04 & 2.11 & \textbf{38.27} & \textbf{12.42} & \textbf{29.48}\\
 & RendBEV  & 70.40 &\textbf{36.18} & \textbf{36.73}& 41.17& \textbf{1.64} & \textbf{5.43} & 36.65& 6.32 & 29.32 \\
 \bottomrule
 \end{tabular}
\end{table*}

\section{Pretraining with GT FV SS}
\label{sec_sup:pretraining_with_gt_ss}
We perform additional experiments by fine-tuning the model obtained with \ourmethod\space using ground truth semantic segmentation labels instead of model predictions on $0.1\%$, $1\%$, $10\%$, $50\%$ and $100\%$ of the training data. We compare the performance of the same architecture pretrained with SkyEye's method and trained from scratch. We report the results in \cref{tab_sup:pretraining-eval-percentages}. We observe that in this setting the model pretrained with \ourmethod\space performs similarly as the one pretrained using model predictions as targets, while SkyEye's results improve by a higher margin. Even with SkyEye's improvement with the usage of GT labels, our method continues to provide better results in low-data regimes, while the difference dissipates in models fine-tuned on 10\% of the data (in the order of 2000 images) and the models pretrained with SkyEye's methodology perform slightly better on higher BEV GT data regimes.

\begin{table*}
\centering
\caption{Impact of the pretraining (with GT PV) on BEV semantic segmentation performance using the network proposed in SkyEye on different data regimes. SkyEye results from \cite{gosala_skyeye_2023}, \ourmethod\space 
 and no pretraining run by us on same splits. All scores are reported on the KITTI-360 dataset.}
\vspace{-0.2cm}
\label{tab_sup:pretraining-eval-percentages}
\scriptsize
\setlength\tabcolsep{3.7pt}
 \begin{tabular}{c|c|cccccccc|c}
 \toprule
 \textbf{BEV GT~(\%)} & \textbf{Pretraining} & \textbf{Road} & \textbf{Sidewalk} & \textbf{Building} & \textbf{Terrain} & \textbf{Person} & \textbf{2-Wheeler} & \textbf{Car} & \textbf{Truck} & \textbf{mIoU} \\
 
 \midrule
\multirow{3}{*}{0.1} & SkyEye & 68.78 & 28.20 & 35.56 & 26.08 & 0.00 & 0.00 & 21.61 & 0.00 & 22.53 \\
& \ourmethod & \textbf{72.15} & \textbf{37.81} & \textbf{36.70} & \textbf{46.65} & \textbf{2.62} & \textbf{3.99} & \textbf{34.56} & \textbf{6.03} & \textbf{30.07} \\
& -- & 56.43 & 19.95 & 23.64 & 7.17 & 0.00 & 0.00 & 12.59 & 0.00 & 14.97 \\

 \midrule
\multirow{3}{*}{1} & SkyEye & 72.56 & 34.33 & 36.70 & 41.66 & 0.00 & 0.16 & 33.85 & \textbf{10.29} & 28.71 \\
& \ourmethod & \textbf{75.33} & \textbf{39.29} & \textbf{38.44}  & \textbf{46.74} & \textbf{3.03}& \textbf{3.95}& \textbf{38.93}  & 8.91 & \textbf{31.82}  \\
& -- & 61.01 & 22.68 & 27.81 & 23.69 & 0.00 & 0.00 & 31.31 & 6.32 & 21.60 \\

 \midrule
\multirow{3}{*}{10} & SkyEye & \textbf{76.07} & 40.30 & 40.30 & 45.33 & 3.75 & \textbf{8.15}  & 42.64 & 10.73 & \textbf{33.41} \\
& \ourmethod & 75.90 & \textbf{40.88} & \textbf{41.06} & \textbf{47.03} & 2.44 & 6.79 & 43.24 & 8.40 & 33.22\\ 
& -- & 73.39 & 37.49 & 35.87 & 40.30 & \textbf{4.72} & 7.44  & \textbf{44.64} & \textbf{12.23} & 32.01 \\ 

\midrule
\multirow{3}{*}{50} & SkyEye & \textbf{76.43} & 39.89 & \textbf{45.22} & 46.64 & \textbf{5.10} & \textbf{7.93}  & 42.43 & 12.30 & \textbf{34.49} \\
& \ourmethod & 74.69 & 40.15 & 42.16 & \textbf{47.22} & 3.30 & 6.78 & 44.88 & 9.77 &  33.61 \\ 
 & -- & 75.30 & \textbf{40.61} & 41.79 & 45.34 & 2.88 & 6.64 & \textbf{45.52} & \textbf{13.46} & \textbf{33.94}  \\ 
\midrule
\multirow{3}{*}{100} & SkyEye & \textbf{75.99} & \textbf{41.35} & \textbf{44.26} & 45.91 & 4.08 & 9.53  & 44.13 & \textbf{12.68} & \textbf{34.74} \\
 & \ourmethod & 75.11 & 40.32 & 42.25 & \textbf{47.55} & 2.91 & 6.89 & 44.19 & 8.51 & 33.47\\
 & -- & 73.01 & 37.78 & 39.15 & 43.68 & \textbf{5.44} & \textbf{10.76} & \textbf{45.41} & 12.25 &  33.72 \\
 \bottomrule
 \end{tabular}

\end{table*}

\section{Additional Qualitative Results}
\begin{figure*}
    \centering
    \includegraphics[width=\linewidth]{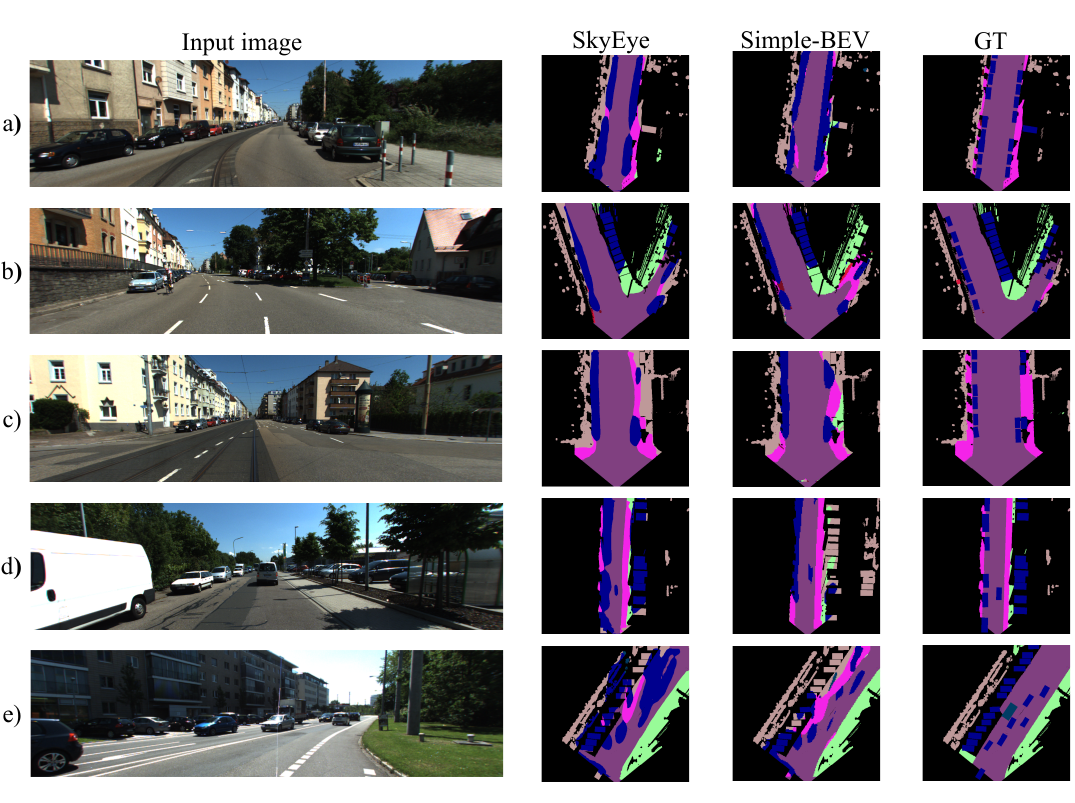}
    \caption{Qualitative comparison of the results obtained with \ourmethod\space using the architectures from SkyEye and Simple-BEV}
    \label{fig_sup:comparison_skyeye_simplebev}
\end{figure*}

We present additional qualitative results from our experiments. In \cref{fig_sup:comparison_skyeye_simplebev} we provide a comparison of the results obtained with the network from SkyEye and Simple-BEV training in a self-supervised way following our method.

\section{Experimental Details}
In this section we provide further details on the experiment configurations and the hardware used to run those experiments.

We feed the BEV network with frames of resolution $1408\times384$, while for Behind the Scenes we resize the images to a resolution of $640\times192$ used in the original paper \cite{wimbauer_behind_2023}. We use a BEV resolution of $768\times704$, which corresponds to a real world area of $56.83m\times52.096m$ in front of the vehicle.

In our experiments, when using a class-weighted cross entropy loss, we use the class weights proposed in \cite{gosala_skyeye_2023}.
When sampling 3D points along rays, we sample in a total of $m=64$ points on each ray with $z_{near}=3 \si{\m}$ and $z_{far}=80 \si{\m}$.

For our self-supervised training, we use a batch of 5 sequences. For each sequence, we sample a total of 192 patches of $16\times16$ pixels randomly distributed across  7 other frames of the sequence, with timestamps $T = \{r-1, r+1, r+o_{1}, \dots, r+o_{5}\}$, where each temporal offset $o_{k}$ is selected in a random uniform way from ranges of length 7 starting from $r+5$. The goal of this selection is to provide a good coverage in different regions of the BEV, as discussed at the end of Sec. 3 and shown in Fig. 3 of the main paper. We train for 20 epochs and use SGD as optimizer with Nesterov momentum $0.9$, weight decay $0.00001$ and learning rate $0.005$.

In terms of hardware, we perform most of our experiments in a machine equipped with a NVidia V100 GPU with 32GB of VRAM. The self-supervised training experiments with 196 patches per sequence take approximately 8 days to complete in a single machine. The neural network architecture proposed in SkyEye \cite{gosala_skyeye_2023}, which we use in our main experiments has 14.6 million parameters and a runtime of 77.84 ms for a forward pass in inference.

\section{Ethical considerations}

In this section we address potential ethical implications of our work. We would like to focus on two main topics: data and possible misuse. 

In terms of data, we don't introduce any new dataset and use for our experiments two publicly available datasets: the KITTI-360 dataset\cite{liao_kitti-360_2023} and the Waymo dataset\cite{waymo_2020_CVPR} as well as their BEV derivations provided by the authors of SkyEye\cite{gosala_skyeye_2023}. The KITTI-360 dataset is shared under a CC BY-NC-SA 3.0 License, while the Waymo dataset is shared under the Waymo Dataset License Agreement for Non-Commercial Use. The BEV version of these datasets are licensed under a non-commercial RL License Agreement. We credit the original authors for the creation of these datasets. For these datasets, appropriate measures (e.g. the blurring of faces and license plates) have been taken in order to respect individual privacy rights: the KITTI-360 dataset is GDPR-compliant and thus provides extensive privacy protection and the Waymo dataset as per their original authors, was modified to protect individuals' privacy.

In terms of potential misuse, we note that the methodology and models described in this work are research artifacts, not intended for their deployment as-is in safety critical applications like autonomous driving given the limitations described in the main paper.

\onecolumn
\twocolumn
\clearpage